\RequirePackage[T1]{fontenc}
\documentclass[conference]{IEEEtran}
\usepackage{cite}
\usepackage{amsmath,amssymb,amsfonts}
\usepackage{graphicx}
\usepackage{textcomp}
\usepackage{xcolor}
\usepackage{iftex}
\ifPDFTeX
\else
  \AtBeginDvi{}
\fi

\begin{document}
\title{EviGround-CT: Sentence-to-Volume Grounding for Verifying LLM-Generated Chest CT Reports}
\author{
\IEEEauthorblockN{Ziyuan Li\textsuperscript{1}, Juncheng Ni\textsuperscript{1}, Feng Li\textsuperscript{1,2 *}}
\IEEEauthorblockA{\textsuperscript{1} Technical University of Munich, Munich, Germany}
\IEEEauthorblockA{\textsuperscript{2} Munich Center for Machine Learning, Munich, Germany}
\IEEEauthorblockA{\textsuperscript{*} Corresponding author. Email: feng.li@tum.de}
}
\maketitle

\begingroup
\renewcommand{\thefootnote}{}
\footnotetext{\textcopyright{} 2026 IEEE. Personal use of this material is permitted. Permission from IEEE must be obtained for all other uses, in any current or future media, including reprinting/republishing this material for advertising or promotional purposes, creating new collective works, for resale or redistribution to servers or lists, or reuse of any copyrighted component of this work in other works.}
\endgroup

\begin{abstract}
Chest computed tomography (CT) reports generated by large language models (LLMs) can describe positive findings that lack localized evidence in the scan. We propose EviGround-CT to verify these findings after report generation. A deterministic extractor identifies positive finding sentences, and a 3D grounding model predicts an evidence mask for each sentence. Using the full sentence retains anatomical location, laterality, and descriptive modifiers in the grounding query. We evaluate report generation on CT-RATE, and grounding and verification on a held-out 100-case ReXGroundingCT subset. Using the available annotations, sentence-grounded verification raises evidence-supported precision (ESP) from 0.263 to 0.365 across 137 extracted in-scope positives, compared with category conditioning. The corresponding unsupported positive finding rate falls from 0.737 to 0.635. The paired case-level bootstrap ESP difference is +0.102 (95\% confidence interval: [0.028, 0.181]). The predicted masks associate individual report statements with spatial evidence that can be inspected alongside the CT.
\end{abstract}

\section{Introduction}
Full-volume chest computed tomography (CT) report generation must summarize hundreds of slices containing abnormalities that may be small, diffuse, or spatially distributed. Recent volumetric datasets and CT-oriented systems enable report generation from these scans~\cite{hamamci2024developing,hamamci2024ct2rep}. Spatial correspondence is also central to multimodal image-guided systems~\cite{li2025robotic,li2026robotic,yueintraoperative}. Evaluating the generated findings requires checking where their evidence lies. For a statement such as ``a small pulmonary nodule is present in the right upper lobe,'' the relevant evidence is a nodule in the specified lobe. Lexical similarity and finding-level agreement assess text or labels without explicitly testing this spatial correspondence. Category agreement, for example, treats mentions of a nodule in different lobes alike. Sentence-to-volume grounding includes the stated location in the evidence query.

EviGround-CT performs this check after report generation. A CT-to-report generator produces a report, a deterministic extractor identifies positive finding sentences, and a sentence-conditioned 3D verifier predicts an evidence mask for each candidate. The verifier receives the full sentence, retaining location, laterality, extent, and appearance descriptions. We train the generator and verifier separately: generation requires compact semantic tokens, whereas grounding requires spatial features for voxel-level prediction.

We evaluate generated findings by matching their categories and predicted masks to reference annotations. Evidence-supported precision (ESP) is the fraction of extracted in-scope findings with annotated spatial support; the unsupported positive finding rate (UPR) is its complement. We assess report generation on CT-RATE and grounding, verification, threshold sensitivity, and ablations on ReXGroundingCT. Comparing category- and sentence-conditioned verifiers on identical reports tests the benefit of sentence context.

\section{Related Work}
\subsection{3D CT Report Generation with LLMs}
CT-RATE, CT2Rep, and 3D-CT-GPT extend image--text learning to volumetric CT~\cite{hamamci2024developing,hamamci2024ct2rep,chen20243d}. These systems encode spatially distributed findings for report generation. Our work locates evidence for the positive findings in their output.

\subsection{Medical Image Modeling and Grounding}
Medical image grounding links textual findings to spatial regions. LViT incorporates medical text into image segmentation~\cite{li2024lvit}, while SAT uses medical terminology as prompts for segmenting anatomical structures and lesions in 3D images~\cite{zhao2025sat}. For chest CT, RadGenome-Chest CT and ReXGroundingCT pair finding descriptions with regions or volumetric masks~\cite{zhang2024radgenome,baharoon2025rexgroundingct}. We train our verifier on ReXGroundingCT sentence--mask pairs and use full generated finding sentences as queries, retaining location, laterality, and descriptive modifiers. The resulting masks associate report statements with candidate evidence regions. Category matching and comparison with reference masks then assess whether each finding has annotated spatial support.

\section{Problem Formulation}
Given a CT volume $V\in\mathbb{R}^{D\times H\times W}$, a generator produces $\hat R=f_{\mathrm{rep}}(V)$. A deterministic extractor selects positive finding candidates $\mathcal S^+\subseteq\mathrm{Sent}(\hat R)$, where $\mathrm{Sent}(\hat R)$ denotes the report sentences. Candidates belong to predefined verifiable categories. For each candidate $s_i$, the verifier predicts
\begin{equation}
\hat M_i=f_{\mathrm{grd}}(V,s_i),\qquad \hat M_i\in[0,1]^{D\times H\times W}.
\end{equation}
The pipeline $V\rightarrow\hat R\rightarrow\mathcal S^+\rightarrow\{\hat M_i\}$ links generated report sentences to volumetric evidence masks. Verification checks category correspondence and spatial support against the reference annotations.

\section{Method}
\begin{figure}[h]
\centering
\includegraphics[width=0.5\textwidth]{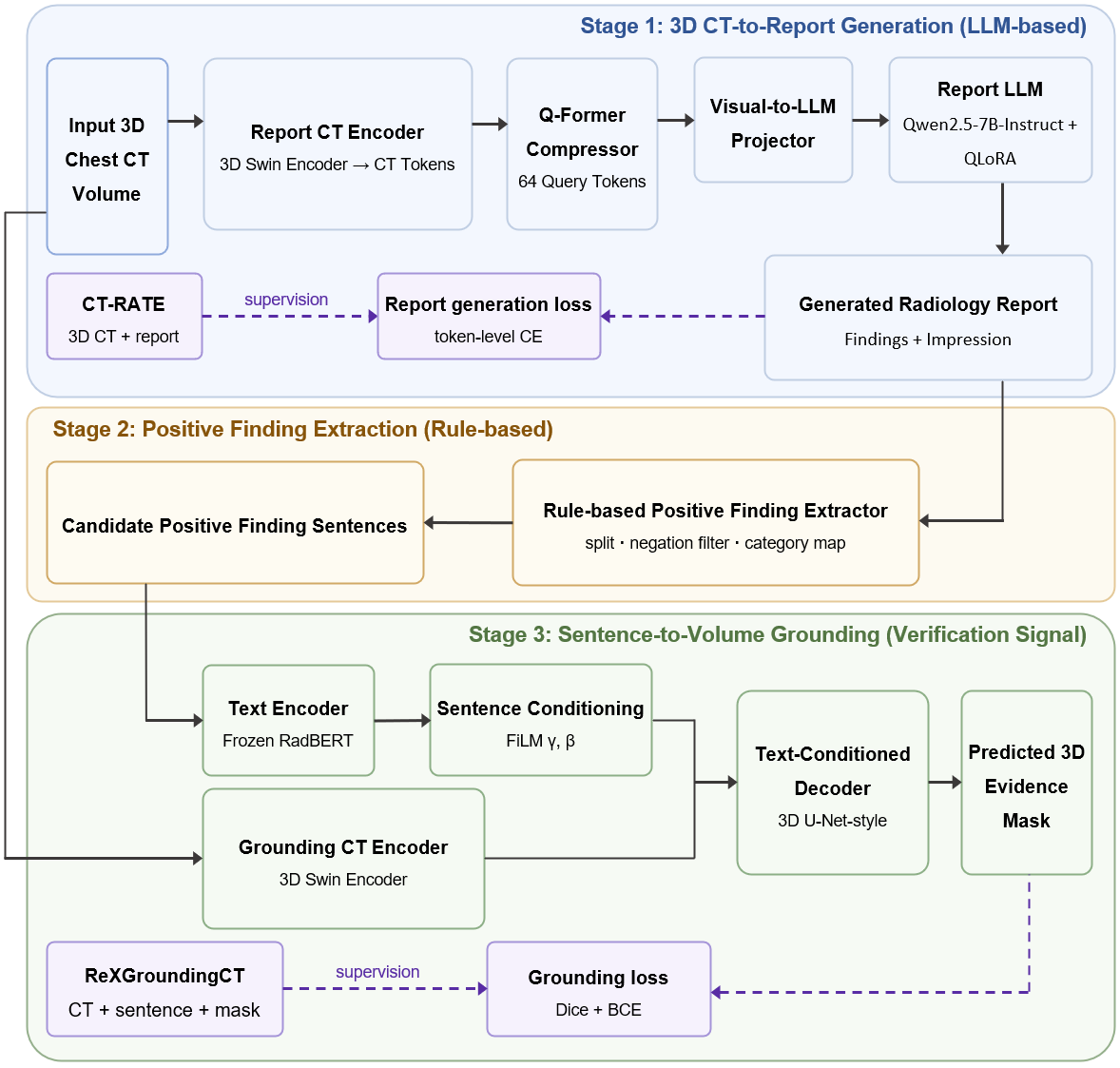}
\caption{EviGround-CT: report generation, positive finding extraction, and sentence-conditioned grounding. Solid arrows indicate inference; dashed arrows indicate supervision.}
\label{fig:overview}
\end{figure}

Figure~\ref{fig:overview} shows the three inference stages. The generator and grounding verifier use separate 3D Swin-style encoders with base widths of 96 and 64 channels, respectively. The narrower grounding encoder reduces volumetric decoding cost. The encoders learn report-token and mask prediction, respectively; at inference, the extractor passes generated finding sentences to the verifier.

\subsection{CT Preprocessing}
We clip single-channel CT volumes to $[-1000,400]$ Hounsfield units (HU), normalize them to $[-1,1]$, and resize them to $96\times192\times192$. Grounding masks undergo the same spatial transformations with nearest-neighbor interpolation.

\subsection{CT-to-Report Generation}
A 3D Swin-style encoder~\cite{liu2021swin} yields hierarchical features $\{F_1,F_2,F_3,F_4\}$. The final map is flattened into tokens $Z\in\mathbb{R}^{N\times C}$, where $N$ is the number of spatial tokens and $C$ is the channel dimension. A Q-Former-style compressor~\cite{li2023blip} uses $K=64$ learnable queries $Q$ to summarize these tokens into a fixed-length sequence:
\begin{equation}
U=\mathrm{CrossAttn}(Q,Z,Z),\qquad H=W_2\sigma(W_1U).
\end{equation}
The queries attend to the CT tokens to produce the compressed sequence $U$. A two-layer projector, with nonlinear activation $\sigma$, maps $U$ to the embedding space of Qwen2.5-7B-Instruct~\cite{qwen2024qwen2}. The resulting soft visual tokens $H$ precede the instruction prompt $I$. With QLoRA~\cite{dettmers2023qlora}, we freeze the quantized LLM backbone and train its adapters jointly with the CT encoder, compressor, and projector. For a target report of $L$ tokens $w_1,\ldots,w_L$, we minimize
\begin{equation}
\mathcal L_{\mathrm{rep}}=-\sum_{t=1}^{L}\log p(w_t\mid w_{<t},H,I).
\end{equation}

\subsection{Positive Finding Extraction}
The extractor applies sentence splitting, keyword matching, negation filtering, and category mapping. Verifiable categories cover nodules, masses, ground-glass opacities, consolidations, atelectasis, pleural effusion, emphysema, fibrosis/scarring, and other opacity-related findings. Negated statements, e.g., ``no pleural effusion,'' are excluded. Unmapped candidates are recorded as out-of-scope and excluded from ESP/UPR. The full retained sentence conditions grounding; the category label is used for reference matching.

\subsection{Sentence-to-Volume Grounding Verifier}
Frozen RadBERT-RoBERTa-4m~\cite{yan2022radbert} encodes each sentence as $e_i=T(s_i)$. Feature-wise linear modulation (FiLM)~\cite{perez2018film} conditions the grounding CT features $F$, and a 3D U-Net-style decoder~\cite{cciccek20163d} predicts the mask:
\begin{equation}
\begin{split}
(\gamma_i,\beta_i)&=\mathrm{MLP}(e_i),\quad \tilde F_i=\gamma_i\odot F+\beta_i,\\
\hat M_i&=D_{\mathrm{grd}}(\tilde F_i).
\end{split}
\end{equation}
The sentence-dependent scale $\gamma_i$ and offset $\beta_i$ modulate CT features before decoding, allowing different queries to produce different masks for the same volume. We train on reference finding sentences paired with ReXGroundingCT masks and use extracted generated sentences for generated-report verification. The grounding loss combines Dice loss and binary cross-entropy (BCE) against the reference mask $M_i$:
\begin{equation}
\mathcal L_{\mathrm{grd}}=\mathcal L_{\mathrm{dice}}(\hat M_i,M_i)+0.5\mathcal L_{\mathrm{bce}}(\hat M_i,M_i).
\end{equation}

\subsection{Evidence-Supported Verification}
Generated findings are processed in report order. Each predicted mask is paired with the unmatched reference mask of the same category having the highest intersection-over-union (IoU), breaking ties by the smallest center distance. The finding has annotated support only if evidence is present ($\geq16$ foreground voxels) and the pair passes either the IoU or center-distance threshold. Only supported matches consume a reference mask; findings without an accepted match remain unsupported. Evidence-supported precision is
\begin{equation}
\mathrm{ESP}=N_{\mathrm{supported}}/N_{\mathrm{gen}},
\end{equation}
where $N_{\mathrm{gen}}$ counts all extracted positives in the predefined verifiable categories, including those without reference matches, and $N_{\mathrm{supported}}$ counts those passing both the category and spatial checks. The unsupported positive finding rate (UPR) is defined as $\mathrm{UPR}=1-\mathrm{ESP}$ and measures the fraction lacking annotated support. Dice and IoU quantify mask overlap after category matching.

\section{Experiments}
\subsection{Datasets}
For CT-RATE~\cite{hamamci2024developing}, we use a case-disjoint subset of 8,000 training, 1,000 validation, and 1,000 test CT--report pairs, preserving abnormality diversity and normal--abnormal balance. ReXGroundingCT~\cite{baharoon2025rexgroundingct} provides 2,992 public training, 50 public validation, and 100 private ReXrank test cases. Since private test masks are unavailable, we reserve 100 training cases for oracle grounding, generated-report verification, threshold sensitivity, and grounding ablations. The verifier is trained on the remaining 2,892 cases, with checkpoints and thresholds selected on the 50 validation cases.

To reduce cross-dataset leakage, CT-RATE training cases matched by case/volume identifiers to ReXGroundingCT validation or held-out evaluation cases are excluded. The held-out subset retains the released, non-exhaustive masks, which provide the reference evidence for evaluation.

\subsection{Implementation Details}
The generator targets Findings and Impression text after metadata removal. Its four-layer compressor has hidden size 768; LoRA rank, alpha, and dropout are 32, 64, and 0.05. Training uses AdamW, bfloat16, gradient checkpointing, accumulation, gradient clipping, and early stopping on validation report loss. Only target tokens contribute to the loss. Inference uses deterministic decoding with at most 512 new tokens.

For the verifier, we train the CT encoder, FiLM module, and decoder with RadBERT frozen. We use the same optimizer family and precision, with early stopping on validation Dice. At inference, probabilities $\geq0.5$ define foreground. Evidence presence requires at least 16 foreground voxels without altering the predicted mask. Training uses two NVIDIA H100 GPUs with random seed 42. Results use the best validation checkpoint, with uncertainty estimated by case-level bootstrap.

\subsection{Baselines and Ablations}
Report baselines are a CT2Rep-style model~\cite{hamamci2024ct2rep} and frozen Qwen2.5 with only the visual projector trained. Grounding baselines use a center/lung prior, Grad-CAM~\cite{selvaraju2017grad}, category conditioning, or frozen RoBERTa-base. Category conditioning uses a finding label; text conditioning uses the full sentence. RoBERTa-base replaces RadBERT in both the oracle comparison and text-encoder ablation, retaining FiLM and the decoder.

Grad-CAM uses the last feature map before global pooling in a multilabel 3D classifier trained on the same categories. Gradients target the category logit. Nonnegative CAMs are trilinearly upsampled to the evaluation grid, normalized to $[0,1]$ per volume and category, and binarized at $>0.5$; maps without positive activation yield empty masks.

Trainable baselines use the same splits, preprocessing, optimizer family, and early stopping where applicable. Ablations remove adapters, visual compression, or text conditioning, replace RadBERT, or share CT encoders. The shared variant uses a single CT encoder with base width 96; the report compressor, grounding FiLM module, and decoder remain task-specific, with dimensions matched to its features.

\subsection{Evaluation}
We evaluate report quality with BLEU-4~\cite{papineni2002bleu}, ROUGE-L~\cite{lin2004rouge}, METEOR~\cite{banerjee2005meteor}, CIDEr~\cite{vedantam2015cider}, and finding F1, using the same rule-based mapper for predefined categories in generated and reference reports. In oracle grounding, each reference sentence is evaluated against its paired mask using Dice, IoU, localization accuracy (fraction passing the spatial criterion), and center-distance error. Dice and IoU use a foreground threshold of $\geq0.5$ with $10^{-6}$ smoothing in numerator and denominator. An empty mask has no foreground voxels; two empty masks score 1, while one empty mask scores near 0. All pairs contribute to mean Dice/IoU. Center distance is the Euclidean distance between foreground centroids in millimeter coordinates, accounting for scan geometry and resizing; its mean includes only pairs with two nonempty masks, while any empty mask fails the spatial support criterion.

For generated reports, Dice and IoU are averaged over findings with a category-matched reference. The default spatial criterion is IoU $\geq 0.10$ or center distance $\leq 30$ mm; relaxed and strict settings are $(0.05, 40\text{ mm})$ and $(0.20, 20\text{ mm})$, respectively. This rule evaluates localization, while Dice and IoU quantify mask overlap. We report 95\% CIs from 1,000 case-level bootstrap resamples and paired ESP differences for identical generated findings.

\section{Results}
\subsection{Report Generation on CT-RATE}
The EviGround-CT generator (Qwen2.5-QLoRA) scores highest on all five report-generation metrics in Table~\ref{tab:report_generation}. Relative to CT2Rep-style generation, BLEU-4 rises from 0.183 to 0.211 and CIDEr from 0.219 to 0.287. Finding F1 is 0.221, compared with 0.158 for CT2Rep-style generation and 0.143 for frozen Qwen2.5. Training the visual pathway and LLM adapters thus yields better agreement with reference findings than training the projector alone.

\begin{table}[h]
\centering
\caption{Report generation on CT-RATE: text similarity and finding F1. Higher is better.}
\label{tab:report_generation}
\footnotesize
\setlength{\tabcolsep}{2pt}
\begin{tabular}{lccccc}
\hline
Method & BLEU-4 & ROUGE-L & METEOR & CIDEr & \shortstack{Finding\\F1} \\
\hline
CT2Rep-style & 0.183 & 0.331 & 0.182 & 0.219 & 0.158\\
Frozen Qwen2.5 & 0.167 & 0.308 & 0.164 & 0.181 & 0.143\\
Qwen2.5-QLoRA & \textbf{0.211} & \textbf{0.359} & \textbf{0.206} & \textbf{0.287} & \textbf{0.221}\\
\hline
\end{tabular}
\end{table}

\subsection{Oracle Sentence Grounding on ReXGroundingCT}
Table~\ref{tab:oracle_grounding} compares grounding of reference findings. This setting fixes the finding descriptions and evaluates the verifier against their paired masks. Category conditioning improves on spatial priors and CAMs, while both sentence-conditioned decoders improve further. RadBERT reaches 0.375 Dice and 0.724 localization accuracy, versus 0.288 and 0.622 for RoBERTa-base; center-distance error falls from 24.9 to 18.2 mm. The consistent gains across overlap and localization metrics support using a radiology-specific sentence encoder.

\begin{table}[h]
\centering
\caption{Oracle grounding of reference findings on the ReXGroundingCT holdout. Loc. acc.: localization accuracy; Dist.: center-distance error.}
\label{tab:oracle_grounding}
\footnotesize
\setlength{\tabcolsep}{3pt}
\begin{tabular}{lcccc}
\hline
Method & Dice $\uparrow$ & IoU $\uparrow$ & Loc. acc. $\uparrow$ & Dist. (mm) $\downarrow$\\
\hline
Center/lung prior & 0.055 & 0.031 & 0.177 & 63.5\\
Grad-CAM & 0.176 & 0.115 & 0.356 & 47.8\\
Category decoder & 0.210 & 0.133 & 0.514 & 31.0\\
RoBERTa-base decoder & 0.288 & 0.179 & 0.622 & 24.9\\
EviGround-CT & \textbf{0.375} & \textbf{0.239} & \textbf{0.724} & \textbf{18.2}\\
\hline
\end{tabular}
\end{table}

\subsection{Verification of Generated Reports}
Table~\ref{tab:generated_verification} reports verification of findings extracted from generated reports. The category and sentence rows use the same Qwen2.5-QLoRA generator and extracted findings; the sentence row is the full EviGround-CT pipeline. CAM denotes auxiliary-classifier Grad-CAM. For Qwen2.5-QLoRA reports, category conditioning improves ESP over CAM from 0.249 to 0.263. Full-sentence conditioning yields a larger gain, reaching 0.365 ESP and reducing UPR from 0.737 to 0.635. The predicted masks also have higher overlap with the references: Dice rises from 0.199 to 0.286, and IoU from 0.123 to 0.169.

\begin{table}[h]
\centering
\caption{Generated-report verification on ReXGroundingCT. Category and sentence conditioning use identical findings.}
\label{tab:generated_verification}
\footnotesize
\setlength{\tabcolsep}{4pt}
\begin{tabular}{lccc}
\hline
Generator + verifier & Dice $\uparrow$ & IoU $\uparrow$ & ESP $\uparrow$\\
\hline
CT2Rep-style + CAM & 0.142 & 0.091 & 0.122\\
Qwen2.5-QLoRA + CAM & 0.159 & 0.104 & 0.249\\
Qwen2.5-QLoRA + category & 0.199 & 0.123 & 0.263\\
Qwen2.5-QLoRA + sentence & \textbf{0.286} & \textbf{0.169} & \textbf{0.365}\\
\hline
\end{tabular}
\end{table}

Of the same 137 generated verifiable positives, 50 pass the default criterion with sentence conditioning and 36 with category conditioning. The paired ESP difference is +0.102 [0.028, 0.181]. Tightening the spatial criterion reduces the supported count for both verifiers (Table~\ref{tab:reliability_analysis}). The paired ESP difference is +0.117 under the relaxed criterion and +0.088 under the strict criterion; all three bootstrap intervals lie above zero. Sentence conditioning therefore retains its advantage as the required spatial agreement becomes stricter.

\begin{table}[h]
\centering
\caption{Spatial threshold sensitivity for 137 findings: ESP and paired differences with 95\% case-level bootstrap CIs.}
\label{tab:reliability_analysis}
\footnotesize
\setlength{\tabcolsep}{4pt}
\begin{tabular}{lccc}
\hline
Criterion & Category & Sentence & $\Delta$ESP\\
\hline
\shortstack[l]{Relaxed\\\strut} & \shortstack{0.328\\{}[0.255, 0.411]} & \shortstack{0.445\\{}[0.365, 0.529]} & \shortstack{+0.117\\{}[0.043, 0.197]}\\[3pt]
\shortstack[l]{Default\\\strut} & \shortstack{0.263\\{}[0.196, 0.342]} & \shortstack{\textbf{0.365}\\{}\textbf{[0.289, 0.448]}} & \shortstack{\textbf{+0.102}\\{}\textbf{[0.028, 0.181]}}\\[3pt]
\shortstack[l]{Strict\\\strut} & \shortstack{0.168\\{}[0.115, 0.239]} & \shortstack{0.255\\{}[0.190, 0.334]} & \shortstack{+0.088\\{}[0.021, 0.164]}\\
\hline
\end{tabular}
\par\smallskip
\parbox{0.96\columnwidth}{Supported counts (category/sentence): 45/61, 36/50, and 23/35 for relaxed, default, and strict criteria. $\Delta$ESP: sentence minus category.}
\end{table}

\subsection{Qualitative Evidence Grounding}
\begin{figure}[h]
\centering
\includegraphics[width=0.5\textwidth]{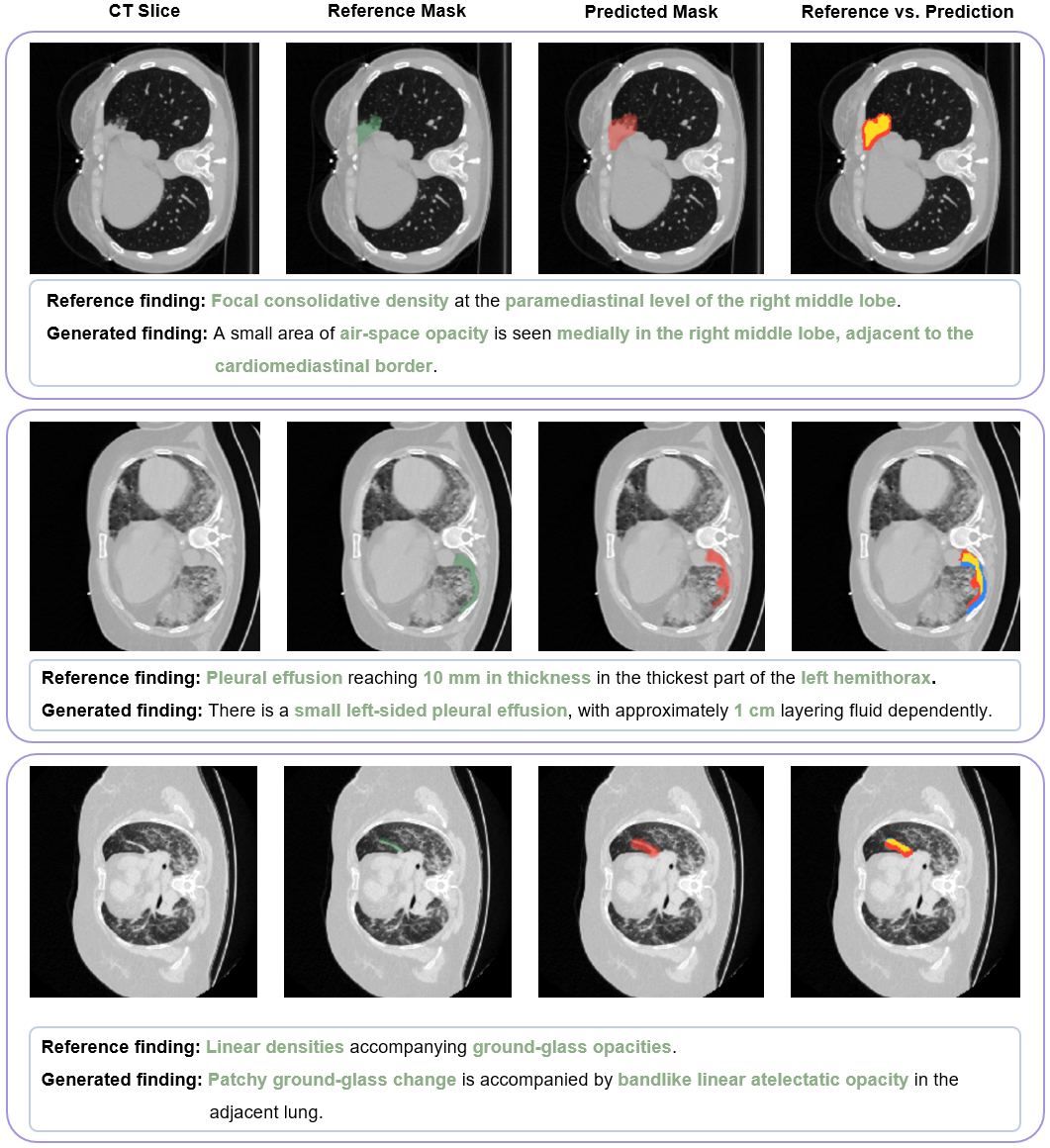}
\caption{Qualitative examples of generated-report verification. Overlay: reference (blue), prediction (red), and overlap (yellow). Highlighted phrases indicate the matched finding and anatomical location descriptions.}
\label{fig:qualitative_results}
\end{figure}

Figure~\ref{fig:qualitative_results} shows right-middle-lobe opacity, left pleural effusion, and ground-glass-related linear opacity. In the top case, the reference describes ``focal consolidative density'' at the ``paramediastinal level of the right middle lobe,'' while the generated sentence describes ``air-space opacity'' medially in the same lobe. The descriptions differ in wording but refer to the same finding and location, and their masks overlap in the corresponding region. The pleural-effusion example preserves left-sided location and approximate thickness: 10 mm in the reference and 1 cm in the generated description. The overlays also show boundary mismatches across the three cases.

\subsection{Ablation Study}
The ablations in Table~\ref{tab:ablation} use the held-out ReXGroundingCT subset, including its reference reports for finding F1; Table~\ref{tab:report_generation} uses CT-RATE. Removing LLM adapters lowers finding F1 from 0.218 to 0.132 and ESP from 0.365 to 0.197. Without visual compression, finding F1 falls to 0.146 and ESP to 0.199. Both changes to the generation branch reduce agreement with the reference reports and the fraction of generated findings with spatial support.

In the grounding branch, removing text conditioning reduces verification Dice from 0.286 to 0.148 and ESP to 0.221. This is the largest drop in both grounding metrics among the verifier ablations. RoBERTa-base recovers some of this loss, reaching 0.240 Dice and 0.303 ESP, but remains below RadBERT. Sharing the CT encoder gives 0.232 Dice and 0.299 ESP. Using separate encoders in the full model improves both mask overlap and the fraction of supported findings.

\begin{table}[h]
\centering
\caption{Generation and verification ablations on ReXGroundingCT. Dice and ESP assess generated-report verification.}
\label{tab:ablation}
\footnotesize
\setlength{\tabcolsep}{4pt}
\begin{tabular}{lccc}
\hline
Variant & Finding F1 & Dice & ESP\\
\hline
Full EviGround-CT & \textbf{0.218} & \textbf{0.286} & \textbf{0.365}\\
\hline
\multicolumn{4}{l}{\textit{Report generation branch}}\\
Without LLM adapters & 0.132 & N/A & 0.197\\
Without Q-Former & 0.146 & N/A & 0.199\\
\hline
\multicolumn{4}{l}{\textit{Grounding verification branch}}\\
Without text conditioning & N/A & 0.148 & 0.221\\
RoBERTa-base encoder & N/A & 0.240 & 0.303\\
Shared CT encoder & N/A & 0.232 & 0.299\\
\hline
\end{tabular}
\par\smallskip
\parbox{0.78\columnwidth}{N/A: metric not reported.}
\end{table}

\section{Discussion}
The category and sentence verifiers receive identical generated findings, allowing a direct comparison of how their queries guide localization. A category label identifies the abnormality type; the sentence also specifies location, laterality, extent, or appearance. The higher ESP and mask overlap with sentence conditioning support retaining these details when predicting evidence regions. Its ESP advantage persists across the tested spatial thresholds.

The generator determines which claims enter verification, and the verifier locates their evidence. Changes to either branch lower ESP in the ablations. Finding F1 and Dice help distinguish changes in report agreement from changes in mask overlap, while the shared-encoder result supports learning separate visual representations for the two tasks. The verifier associates each extracted finding with a predicted region for inspection alongside the report.

ESP and UPR measure support against the available annotations. The holdout masks are non-exhaustive, so UPR is not a clinical hallucination rate. External cohorts and radiologist review would help assess findings omitted from the annotations and extend evaluation to negative and uncertain statements.

\section{Conclusion}
EviGround-CT combines CT-to-report generation, positive finding extraction, and sentence-to-volume grounding to predict evidence masks for generated chest CT findings. On the held-out ReXGroundingCT subset, sentence conditioning raises ESP from 0.263 to 0.365 over category conditioning on the same 137 generated findings, with gains across all tested spatial thresholds. The predicted masks connect individual report statements to candidate evidence regions that can be inspected alongside the scan.

\bibliographystyle{IEEEtran}
\bibliography{reference}
\end{document}